\documentclass[runningheads]{llncs}
\usepackage{graphicx}
\usepackage{amsmath}
\usepackage{amssymb}
\usepackage[sort&compress,numbers]{natbib}
\usepackage{subfigure}

\begin{document}
\title{Model Extraction and Adversarial Attacks on Neural Networks using Switching Power Information}
\titlerunning{Adversarial Attacks using Switching Power}
\author{Tommy Li\inst{1} \and
Cory Merkel\inst{1}}
\authorrunning{T. Li and C. Merkel}
\institute{Brain Lab, Rochester Institute of Technology, Rochester NY 14623, USA
\email{\{txl2747,cemeec\}@.rit.edu}\\
\url{www.rit.edu/brainlab}}
\maketitle              % typeset the header of the contribution

\begin{abstract}
Artificial neural networks (ANNs) have gained significant popularity in the last decade for solving narrow AI problems in domains such as healthcare, transportation, and defense. As ANNs become more ubiquitous, it is imperative to understand their associated safety, security, and privacy vulnerabilities. Recently, it has been shown that ANNs are susceptible to a number of adversarial evasion attacks - inputs that cause the ANN to make high-confidence misclassifications despite being almost indistinguishable from the data used to train and test the network.  This work explores to what degree finding these examples may be aided by using side-channel information, specifically switching power consumption, of hardware implementations of ANNs.  A black-box threat scenario is assumed, where an attacker has access to the ANN hardware’s input, outputs, and topology, but the trained model parameters are unknown.  Then, a surrogate model is trained to have similar functional (i.e. input-output mapping) and switching power characteristics as the oracle (black-box) model.  Our results indicate that the inclusion of power consumption data increases the fidelity of the model extraction by up to 30\% based on a mean square error comparison of the oracle and surrogate weights.  However, transferability of adversarial examples from the surrogate to the oracle model was not significantly affected.

\end{abstract}

\section{Introduction}
\label{intro}
Artificial neural networks (ANNs) have become increasingly popular in the past several years due to a convergence of better training algorithms, faster hardware, and the availability of large labeled datasets.  However, as they become more ubiquitous, ANNs are facing mounting challenges related to their privacy, security, and safety.  In large part, this is due to recent demonstrations that show ANNs such as deep convolutional neural networks (CNNs) can easily be fooled into providing high-confidence misclassifications through small, imperceptibly-perturbed versions of their inputs (a.k.a. adversarial examples) \cite{szegedy2013intriguing}.  The study of these types of issues from a more general machine learning (ML) context (\textit{adversarial machine learning or AML}) can be traced back to the mid-2000`s \cite{dalvi2004adversarial,lowd2005adversarial}.  Today, AML research focus has been amplified by the popularity of deep learning, with over 3000 papers published on AML attacks, defenses, and theory since 2014 alone \cite{amlpapers}.

An important subset of AML research deals with so-called black-box attacks of ML models, where an attacker has no knowledge of the model parameters, but can query the model by controlling its inputs and observing its outputs (e.g. classification).  Through this process, the attacker may be able to learn the model's behavior, or even its exact parameter values, which could hold private or proprietary information.  Furthermore, if the behavior is extracted, then one may craft adversarial examples that cause the model to behave in an unintended way.  In this work, we consider the case where attackers make use of not only model outputs, but also side-channel information, to perform black-box attacks.  side-channel information can be described as unintended or non-primary sources of information about a computation that typically depend on low-level implementation details.  Examples include power consumption, analyzing timing between inputs and outputs, observing emitted sound, and checking memory accesses \cite{sc_def}.  In this work, we focus on power consumption as a source of side-channel information.  A few existing works have explored the ability to extract information about black-box ANN models by measuring power consumption.  Wei et al. utilized a hardware-based ANN's power to recover the inputs to the network \cite{sc_trace}. Yoshida et al. mounted a model extraction attack on a small multilayer perceptron (MLP) model (20 model parameters) implemented on a field programmable gate array (FPGA) using correlation power analysis \cite{model_fpga}. Hua et al. successfully performed a model extraction attack on a CNN by observing read and write memory accesses to extract layer parameters \cite{cnn_extract}. Batina et al. extracted all parameters of an MLP model using timing and power side-channel information \cite{sc_net}. The activation function was recovered using timing analysis, the weights were calculated using correlation power analysis, and the layer parameters were obtained using simple power analysis.

This work expands on these previous studies and provides the following novel contributions:
\begin{itemize}
    \item A Siamese ANN-based methodology for extracting black-box ANN parameters using switching power consumption
    \item A study of the transferability of adversarial examples from the extracted model to the black-box model
\end{itemize}

The rest of this paper is organized as follows: Section \ref{section2} provides necessary background on AML.  Section \ref{section3} details the simulation setup, including ANN parameters, power consumption model, and important metrics.  Section \ref{section5} provides simulation results and analyses, and Section \ref{conclusion} concludes this work.

\section{Background}
\label{section2}

AML concerns both the offensive and defensive measures associated with malperformance and/or privacy of ML.  This paper focuses on offensive measures, or attacks, which can be placed into three categories \cite{joseph2018adversarial,vorobeychik2018adversarial,biggio2018wild}.  1.)  Evasion attacks exploit the idea that most ML models such as ANNs learn small-margin decision boundaries.  Legitimate inputs to the model are perturbed just enough to move them to a different decision region in the input space.  2.)  Poisoning attacks typically use modified labeling or addition of training data to reduce the margins of decision boundaries or insert new boundaries that cause misclassifications and also make evasion attacks easier to perform.  3.)  A third type of attack targets the privacy of ML models and/or training data.  By querying models, these attacks can use statistical methods to infer private information about the parameters of the model or the training set itself.  Of these types, evasion attacks are the most well-studied, especially in deep learning models.  In the mid-2000's, evasion attacks were introduced as small perturbations to the content of emails, causing them to be misclassified by linear spam filters \cite{dalvi2004adversarial,lowd2005adversarial,lowd2005good}.  In 2014, Szegedy et al. \cite{szegedy2013intriguing} showed that imperceptible perturbations in the pixel space of images led to high-confidence misclassifications by CNNs.  The goal of an evasion attack can be expressed as an optimization problem, where, for some model $\Pi$, a correctly-classified input $\mathbf{u}$, usually from the test or training set, is perturbed by $\mathbf{r}^{*}$ to maximize a loss function $\mathcal{L}$ and cause $\Pi$'s classification of $\mathbf{u}^{\prime}=\mathbf{u}+\mathbf{r}^{*}$ to be different from $\mathbf{u}$'s ground truth label:

\begin{equation}
\begin{aligned}
\mathbf{r}^{*}= \underset{\mathbf{r}\in\mathcal{R},}{\arg\max}&\quad \mathcal{L}_{\Pi}(\mathbf{u}+\mathbf{r},l)\\
\textrm{s.t.}&\quad l^{\prime}\ne l
\label{eqn:evasion}
\end{aligned}
\end{equation}

\noindent where $l$ is $\mathbf{u}$'s ground truth label, $l^{\prime}$ is the model's label for $\mathbf{u}^{\prime}$, and $\mathcal{R}$ is the set of set of allowed perturbations.  $\mathbf{u}^{\prime}$ is called an adversarial example.  Often, the allowed set of perturbations takes the form of an $\ell^{p}$-norm constraint: $\mathcal{R}=\{\mathbf{r}\in\mathbb{R}^{N}:\:||\mathbf{r}||_{p}\le D\}$ where $N$ is the dimension of the input space. The $\ell^{p}$-norm of $\mathbf{r}$ is usually bounded in a way that the difference between  $\mathbf{u}$ and $\mathbf{u}^{\prime}$ is difficult or impossible to perceive by a human.  Attacks are usually performed using $\ell^{p}$-norms with $p=2$ or $p=\infty$.  However, $p=0$ and $p=1$ are also common.  $\mathcal{R}$ may also be formed using multiple $\ell^{p}$-norms, box constraints (e.g. bounding all inputs between a minimum and maximum value), or by choosing $\mathbf{r}$ as some type of transformation that imposes a dependence between the elements of $\mathbf{r}$ (e.g. affine transformations such as rotation, scaling, etc.).  A number of evasion attacks have been proposed based on (\ref{eqn:evasion}), differing primarily in the way they define $\mathcal{R}$, how much information they assume is known about $\Pi$ (white-box vs. black-box attacks), the way they approach the optimization procedure, and whether they are targeted (e.g. classifying a school bus image as an ostrich) or untargeted (e.g. classifying a school bus as anything other than a school bus).

This paper focuses on the untargeted fast gradient sign method (FGSM) evasion attack applied to ANN models.  Introduced by Goodfellow et al. \cite{fgsm}, FGSM can be written as
\begin{equation}
\mathbf{r}^{*}=\epsilon\times\mathrm{sgn}\left(\nabla_\mathbf{u}\mathcal{L}_{\Pi}(\mathbf{u},l) \right)
\end{equation}
\noindent
where $\mathrm{sgn}(\cdot)$ is the sign function.  This attack is easy to apply when full details of $\Pi$ (i.e. structure and parameters) are known.  However, a more realistic attack scenario is that limited information about $\Pi$ is available to the attacker.  In this case, $\Pi$ is considered a black-box model, and attacking $\Pi$ is called a black-box attack\footnote{Note that sometimes gray-box is used to describe the situation where some, but not all details of $\Pi$ are known, but here we use the term black-box  to indicate imperfect knowledge of $\Pi$.}.  Specifically, in this work, we assume that the attacker does not know the weights and biases of the black-box ANN model, but does know the ANN topology, activation functions, etc.  We believe that this is a likely scenario, since many applications employ well-known ANNs (e.g. CNNs such as ResNet-50, VGG-16, etc.) and then train them or fine tune them for their particular dataset.  One of the popular methods for performing black-box attacks on ANNs is to estimate their behavior using a surrogate model $\hat{\Pi}$.  Then, adversarial examples can easily be generated for the surrogate model since all of the model details are known (white-box attack).  Finally, the adversarial examples generated against the surrogate model can be transferred to the black-box model.  Note,  in this context, the black-box model is often referred to as the oracle model.  We can define the transferability as the probability  that the oracle's label will be modified by the adversarial example given that the example also modified the surrogate's label and the original labels of both models matched the ground truth target label $l_{t}$:
\begin{equation}
Tr=\mathrm{Pr}\left(l^{\prime}\ne l \left\vert \hat{l}^{\prime}\ne \hat{l} \land  \hat{l}=l=l_{t}\right.\right)
\end{equation}
\noindent
where $l^{\prime}$ and $l$ are the oracle's label of the adversarial and original inputs and $\hat{l}^{\prime}$ and $\hat{l}$ are the surrogate's label of the adversarial and original inputs. In general, the transferability depends on how well the oracle model is extracted and estimated by the surrogate.  Interestingly, transferability does not necessarily depend on $\Pi$ and $\hat{\Pi}$ having identical parameters, and extracting the model behavior is generally much easier than finding the exact parameter values \cite{model_extract,model_data,bb_ml}.  Here, we adopt a query-based approach \cite{bb_ml}, where the surrogate model is trained on examples from the oracle's training set.  The surrogate's target for each input is the label that the oracle assigns to it.  The goal of this work is to determine if additional information (power consumption) from the query will lead to better transferability between the surrogate and the oracle with the same number of queries.

\section{Simulation Setup}
\label{section3}

In this work, the oracle and surrogate models are MLPs with 784 inputs, a single 100-neuron hidden layer, and a 10-neuron softmax output layer, trained on the MNIST dataset \cite{mnist}.  The hidden neurons use a binary activation function:
\begin{equation}
    b(s) = 
     \left\{
    \begin{array}{ll}
          0 & s < 0 \\
          1 & s \geq 0
    \end{array} 
    \right. 
    \label{eq:binary}
\end{equation}
Quantized activations, especially binary activations, are attractive from a hardware perspective, since they require fewer hardware resources (i.e. transistors) and they often have a limited effect on the accuracy of an ANN \cite{yang2019quantization}.  One challenge of training ANNs with binary activations is that the gradient is undefined when the activation function input is 0.  To overcome this challenge, we approximate the gradient as if $b$ were a sigmoid function:
\begin{equation}
    \frac{\partial b(s)}{\partial s}\approx \frac{\partial \sigma(2s)}{\partial s}=\sigma(2s)\left(1-\sigma(2s)\right)
\end{equation}
\noindent
where $\sigma(\cdot)$ is the logistic sigmoid function.  Here, we empirically found that scaling the sigmoid argument by 2 leads to better training results.

\begin{figure}[!t]
\centering
\subfigure[]{
\includegraphics[scale=0.7]{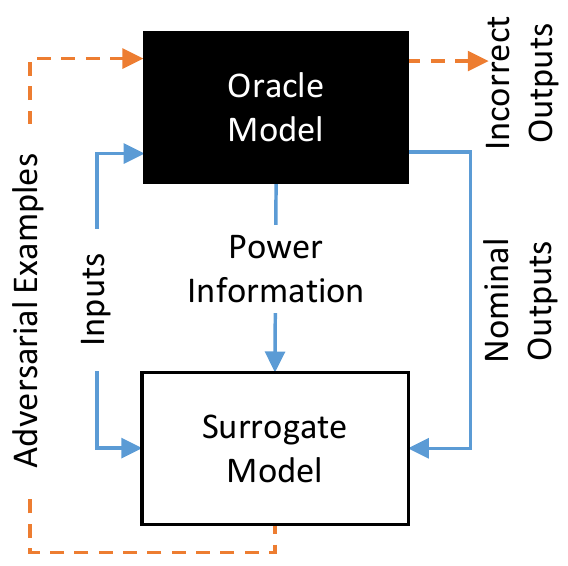}
}
\hspace{10mm}
\subfigure[]{
\includegraphics[scale=1.5]{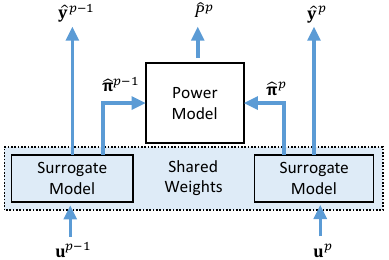}
}
\caption{Extraction of the oracle model using switching power information.  (a)  The surrogate model is trained using both the model outputs (i.e. classification label) and the oracle's switching power.  (b)  A Siamese network structure with shared weights is used to train the surrogate to match the oracle's classifications as well as its switching power.}
\label{fig:power}
\end{figure}

Both the oracle and surrogate models were implemented in tensorflow.  The oracle model was trained on all 60,000 images of the MNIST training set and achieved an average test accuracy of $\approx$90\%.  The surrogate model was trained on different subsets of the MNIST training set, using the outputs of the trained oracle as the target labels.  In addition, the relative switching power of the oracle model's hidden layer was simulated to use as an additional training target for the surrogate, as shown in Figure \ref{fig:power}(a).  Here, we assume that the oracle model is implemented on digital hardware such as an FPGA, and that we can isolate the switching power of the hidden layer neurons from the rest of the power consumption profile.  For a digital circuit, the switching power at a node can be written as
\begin{equation}
P_{switch}=\alpha CV_{dd}^{2}f
\end{equation}
\noindent
where $\alpha$ is the probability that a circuit node changes from 0 to 1 within a clock period, $C$ is the node's capacitance, $V_{dd}$ is the supply voltage, and $f$ is the clock frequency \cite{westecircuits}.  Since $C$, $V_{dd}$, and $f$ are the same for the output of each hidden layer neuron, we can capture their relative power consumption by how often they switch from 0 to 1.  However, note that an attacker would likely only have access to a total, aggregated power profile.  Even if the hidden layer's switching power can be isolated from other power components, the attacker will only know the total switching power.  Therefore, in essence, we can simplify our assumptions by stating that, for each subsequent pair of inputs, the attacker will be able to determine from the power profile the total number of hidden layer neurons that switched from 0 to 1.  Therefore, we redefine the power consumption for a particular input pair as
\begin{equation}
P^{p}=(\mathbf{\pi}^{p}-\mathbf{\pi}^{p-1})(\mathbf{\pi}^{p})^{\top}
\end{equation}
\noindent
where $\mathbf{\pi}$ is a binary vector representing the state (neuron outputs) of the oracle's hidden layer.  Note, that this is just the sum of the number of hidden nodes that switched from 0 to 1 when input $p-1$ switched to input $p$.  Now, the surrogate can be trained using both the oracle outputs and power information, as shown in Figure \ref{fig:power}(b).  Here, we adopt a Siamese network structure, where two surrogate models with shared weights are trained on two inputs that were subsequently used to query the oracle.  The loss function for the surrogate can be written as
\begin{equation}
\mathcal{L}=\mathcal{L}_{CE}(\mathbf{y}^{p-1},\hat{\mathbf{y}}^{p-1})+\mathcal{L}_{CE}(\mathbf{y}^{p},\hat{\mathbf{y}}^{p})+\beta\left[P^{p}-\hat{P}^{p}\right]^{2}
\label{eqn:losspower}
\end{equation}
\noindent
where $\mathcal{L}_{CE}$ is the cross entropy loss, $\beta\ge 0$ is the relative loss of the power consumption, and $\mathbf{y}$ and $\hat{\mathbf{y}}$ or the outputs of the oracle and surrogate models, respectively.

\section{Results and Analysis}
\label {section5}

\subsection{Model Estimation}

\begin{figure}[h]
\centering
\includegraphics[width=0.6\textwidth]{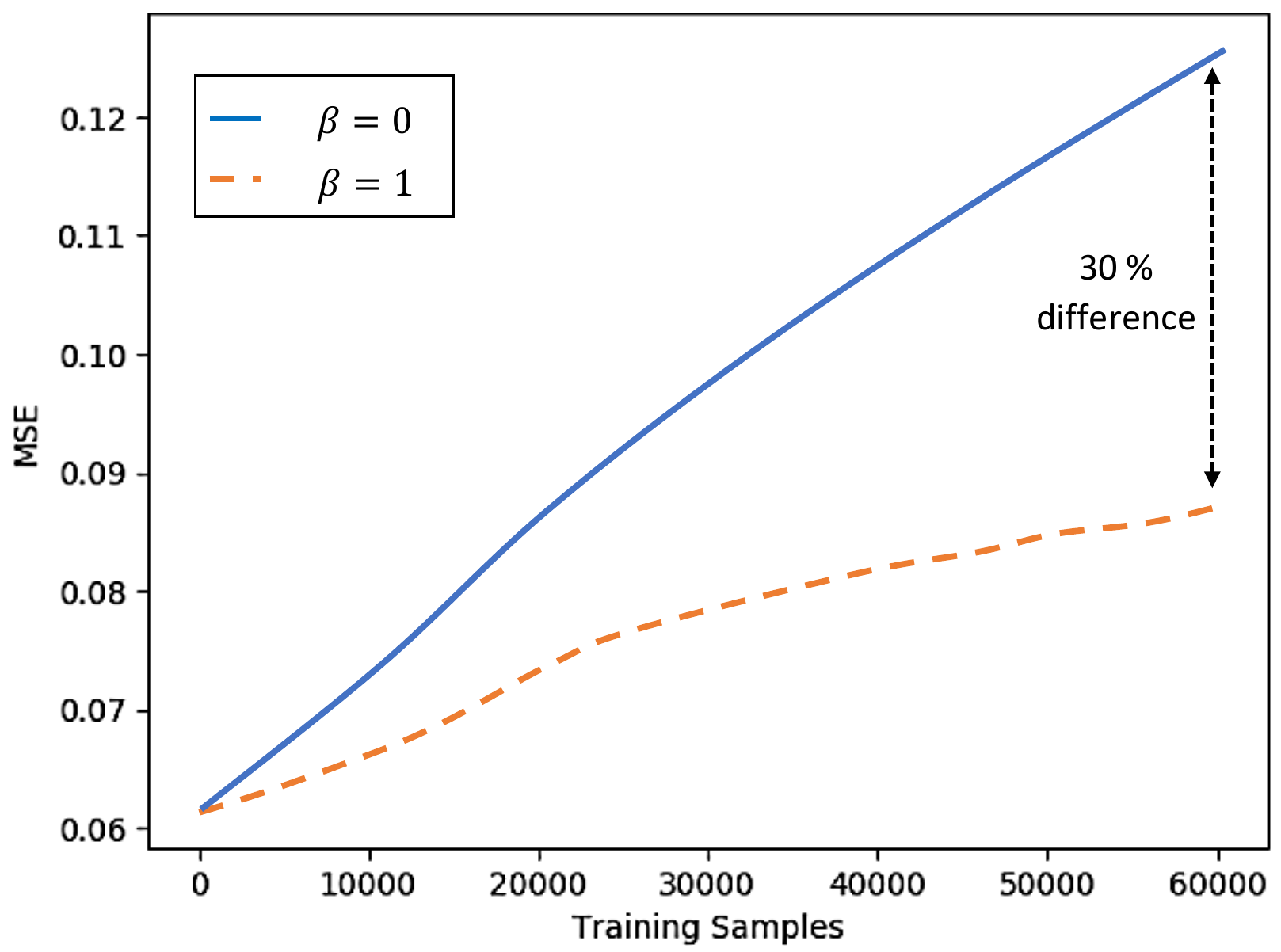}
\caption{MSE of the weights between the oracle and the surrogate models with $\beta=0$ and $\beta=1$.}
\label{fig:mse_mlp}
\end{figure}

In our first set of simulations, we analyzed the effect of including switching power consumption on the efficacy of query-based model estimation.  The surrogate model was trained with different-sized subsets of the MNIST training data, raining from 1 to 60,000. The surrogate model was identically initialized for each training set size. After simulating each training set size, the weights between the oracle and the surrogate were compared using mean-squared error (MSE).  The surrogate was trained on each of the training set sizes 50 times, and the results were averaged. A run consisted of the training of the oracle and each training set size for the surrogate. While the weights before the training of the surrogate were identically initialized during each run, they were not identically initialized between runs. Figure \ref{fig:mse_mlp} shows the relationship between the number of training samples used to train the surrogate and the MSE of the weight matrices.

The MSEs between the various training sample numbers used to train the oracle grows linearly as the number of training examples increases. Initially, this may seem counter-intuitive.  However, there are several valid solutions to the MNIST classification problem within the MLP's weight space. As a result, we can loosely think of the training process for both the oracle and surrogate models as two independent random walks starting at the same point.  From this view, it is clear that their average distance and MSE will grow with more steps.  We also observe that, with the introduction of power information, there is a decrease in MSE for the larger training set sizes - the MSE at 60,000 training samples decreased from 0.13 to about 0.087, or an overall decrease of 30\%. The power information in the loss function likely constrained the weight updates at larger sample sizes.

\subsection{Adversarial Transferability} \label{at}

Next, we studied FGSM attacks against the oracle and surrogate models. First, a white-box attack was performed on the trained oracle using the 10,000 test samples to obtain a set of adversarial images. A white-box attack was also performed on the trained surrogates, regardless of the number of training samples used to train it. Two relative accuracies were calculated - one for the white-box attack on the oracle, and a black-box attack on the oracle, where the adversarial images from the surrogate were used to attack the oracle. A relative accuracy is defined as the accuracy of the model on the adversarial examples divided by the accuracy of the model on the unperturbed images. This metric allows for the comparison of how strong the black-box attack is compared to the white-box attack.
    
\begin{figure}[!t]
\centering
\subfigure[]{
\includegraphics[width=0.75\textwidth]{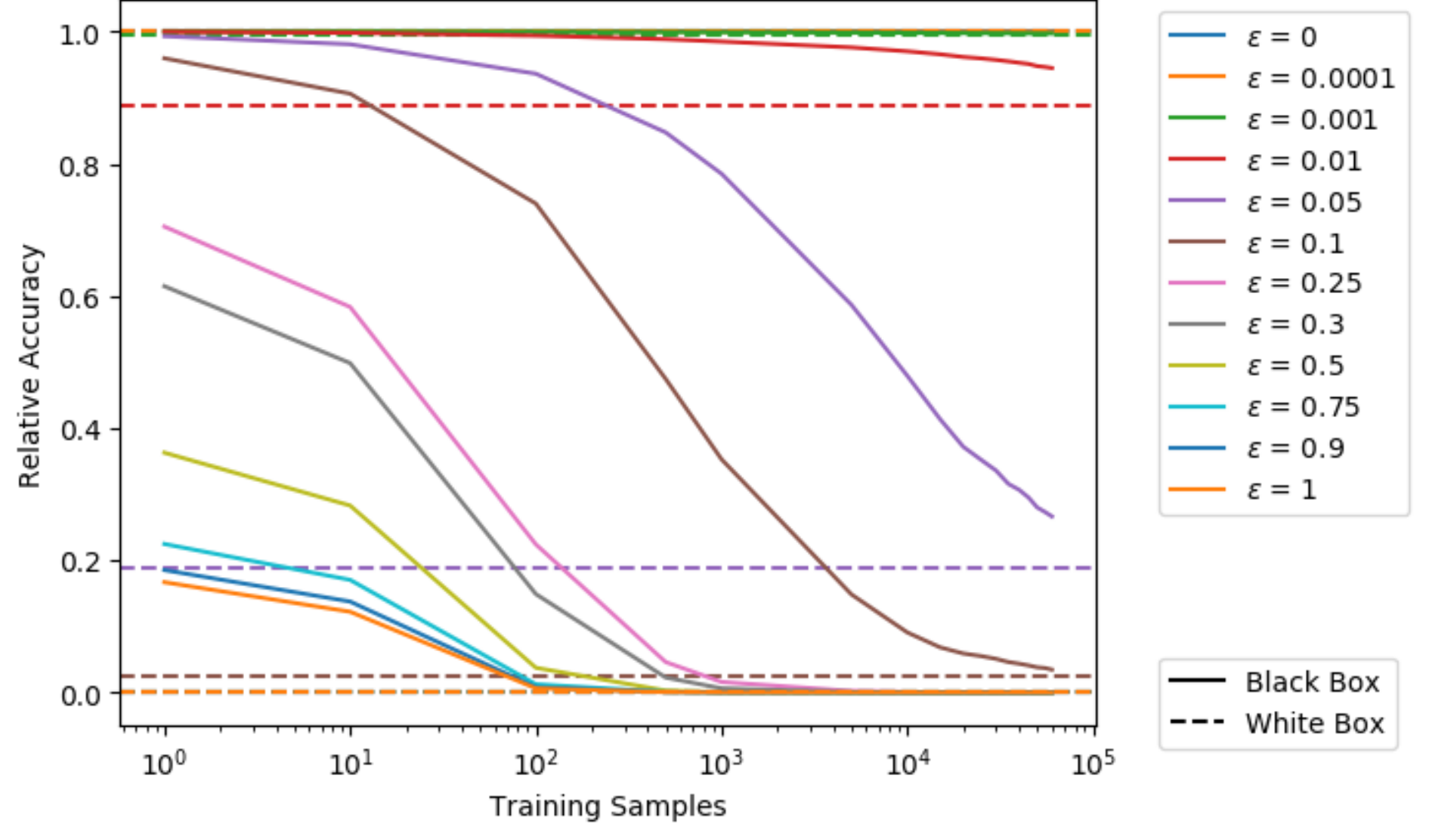}
\label{fig:mse_no_power}
}
\subfigure[]{
\includegraphics[width=0.75\textwidth]{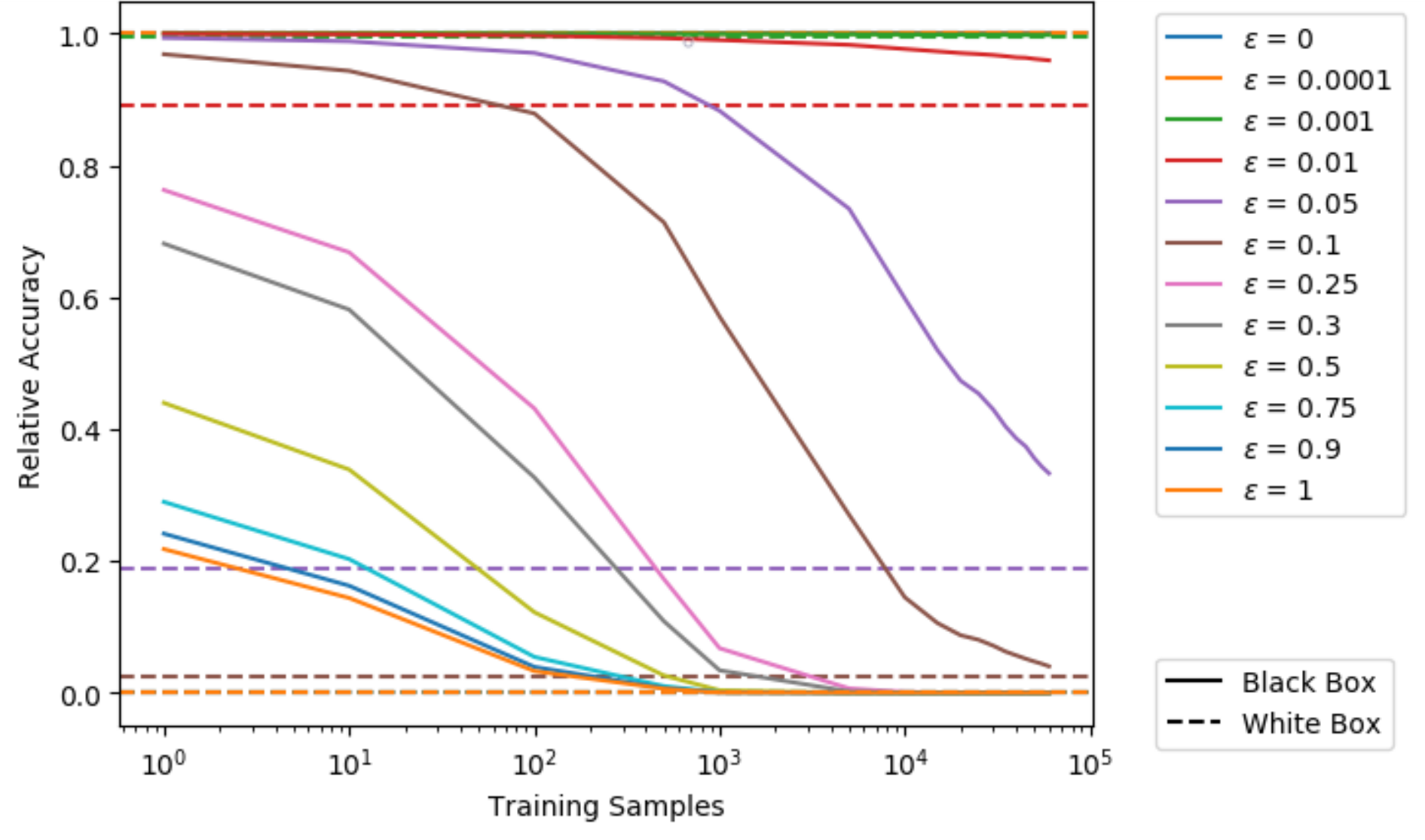}
\label{fig:mse_power}
}
\caption{White-box vs. black-box relative accuracies for (a) $\beta=0$ and (b) $\beta=1$.}
\label{fig:rel_acc}
\end{figure}

Several values for the strength, $\epsilon$, were used to test the effects of the scaling on the attack. Values used for $\epsilon$ ranged from 0 to 1. More $\epsilon$ values that were tested were between 0.1 and 1, as lower $\epsilon$ values did not add enough noise to the adversarial image to cause a large number of misclassifications.  Figure \ref{fig:rel_acc} shows the average relative accuracy plot of the white-box attack on the oracle vs. the black-box attack on the oracle. Relative accuracy is defined as the accuracy of adversarial images divided by accuracy of unperturbed images.  At lower $\epsilon$ values, the relative accuracies for both sets of attacks are very close to 1.0, as not much noise was added to the image. The first noticeable change in relative accuracy occurs at $\epsilon$ = 0.01. For all $\epsilon$, the relative accuracy from the black-box attack asymptotically approaches the relative accuracy from the white-box attack. This is expected, as the differing weights between the oracle and surrogate would produce different gradients, and thus, different perturbations would be generated in the attack. For the attacks with the power information, the overall relative accuracy of the black-box attack was higher (such as 0.01 and 0.05), which implies the networks are more resistant to the adversarial examples.  We believe that this is likely due to additional and unintended regularization of the surrogate model coming from the power loss.  
    
%Next, the transferability of the attack was considered. The transferability of the attack is defined as the percentage of attacks that were successful on the surrogate that were also successful on the oracle:
    
%\begin{equation}
%    T = \frac{A_{oracle} - A_{0,oracle}}{A_{surr} - A_{0,surr}}
%    \label{eq:mse}
%\end{equation}

%\noindent where $A_{oracle}$ and $A_{surr}$ are the number of adversarial samples that caused a misclassification on the oracle and surrogate, respectively, and $A_{0,oracle}$ and $A_{0,surr}$ are the number of samples misclassified that were also misclassified without perturbation. For both cases, attacks only count for the transferability metric if the unperturbed sample was correctly classified in both the oracle and surrogate (or at $\epsilon$ = 0). Figure \ref{fig:transfer} shows the transferability of attacks from the surrogate to the oracle.

\begin{figure}[!t]
\centering
\subfigure[]{
\includegraphics[width=0.75\textwidth]{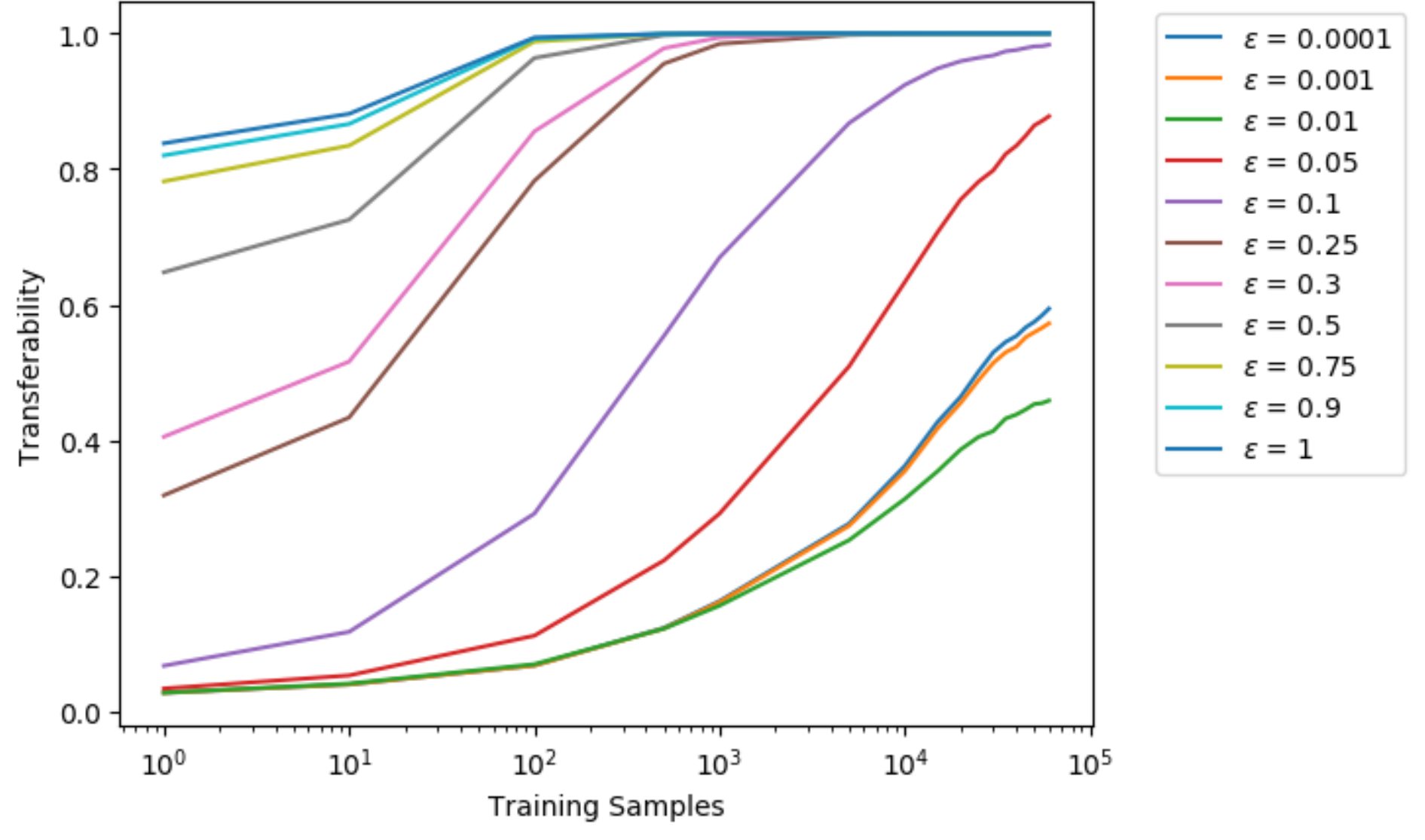}
}
\subfigure[]{
\includegraphics[width=0.75\textwidth]{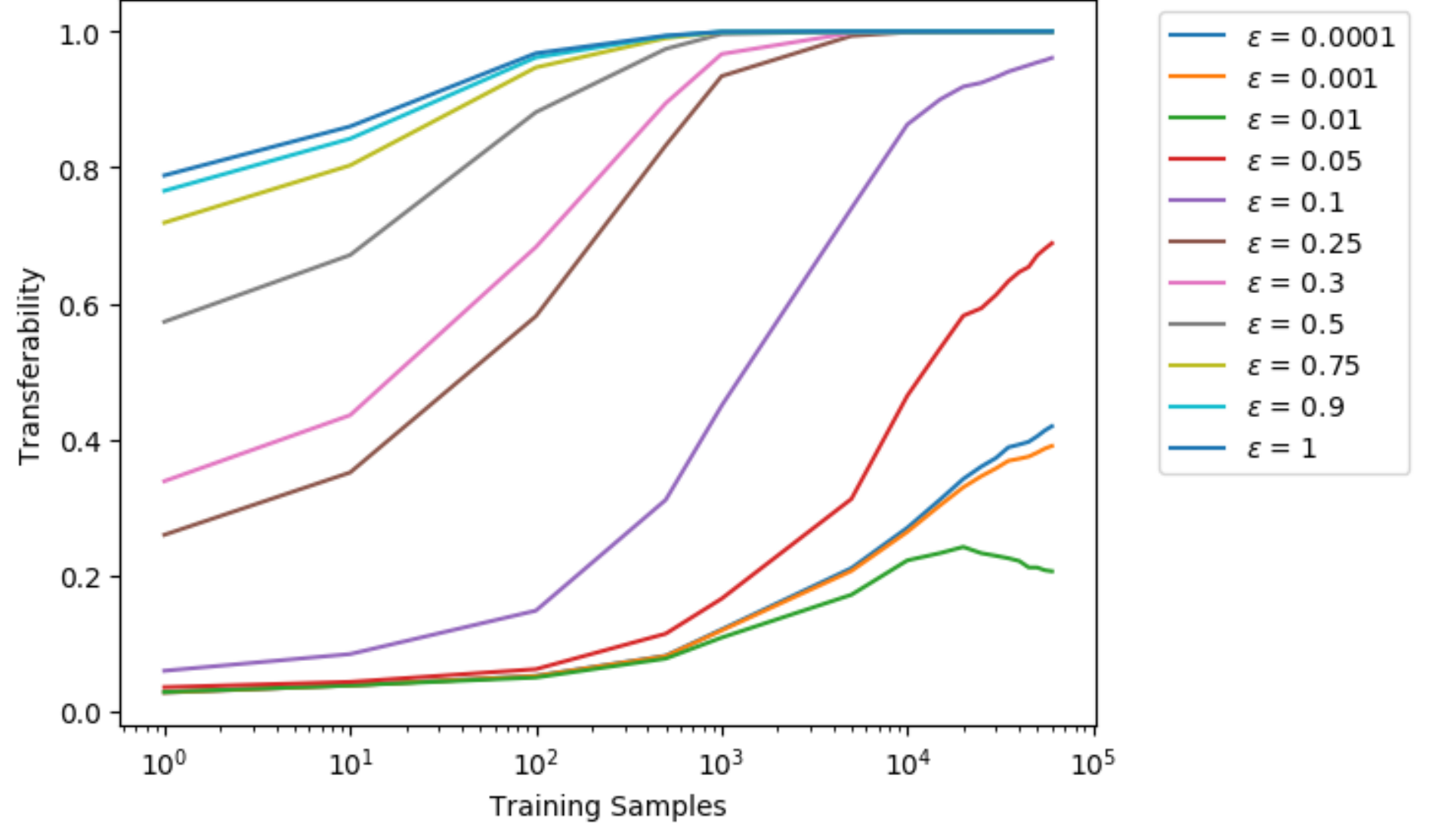}
}
\caption{Transferability of the FGSM attack for (a) $\beta=0$ and (b) $\beta=1$.}
\label{fig:transfer}
\end{figure}

The transferability of the attacks are presented in Figure \ref{fig:transfer}.  At lower $\epsilon$ values, attacks are less likely to transfer, as the adversarial images are unlikely to cause either the oracle or surrogate to mispredict. As expected, more training samples result in higher transferability, as the functionality between the two networks is closer. As $\epsilon$ increases, surrogates trained on fewer training examples are more likely to have attacks transferred, as there was increased noise being added to the image. With power information, the overall transferability remained approximately the same and sometimes even decreased.  One potential reason for this is that we observed the power component of the loss function often reached a local minimum.  Since there are so many possible hidden layer switching behaviors that would lead to the same sum of total switches for two subsequent inputs, it is likely that the power loss landscape is highly non-convex.  Further investigation into techniques to better optimize the power loss will be needed in order for this approach to improve transferability.

\section{Conclusions}
\label {conclusion}

This work explored the use of power consumption information as a means to improve the query efficiency of surrogate-based black-box attacks on artificial neural networks.  Our results indicate that including switching power information in the training of the surrogate model can lead to a significant improvement in the fidelity of model extraction (up to 30\%) as measured by the MSE of the surrogate and oracle weights.  However, we did not observe a significant change in the transferability of attacks from the surrogate to the oracle when power consumption data was included.  This is likely due to the idea that the power loss is highly non-convex, and likely settles into a local minimum.  Future directions for this work may include the exploration of other optimization techniques, such as genetic algorithms for minimizing the power loss, or smarter querying that allows more efficient integration of the power data into the surrogate training process.

%\section{Improving the Usefulness of side-channel Information}

%In this paper, we have assumed that the power consumption of the neurons in each layer of the network can be separated from the rest of the power profile.  The dynamic power consumed is proportional to the number of neurons that switch from 0 to 1:
%\begin{equation}
%\begin{split}
%\hat{P}=&\left(\mathbf{x}_{t}-\mathbf{x}_{t-1}\right)\mathbf{x}_{t}^{\top}\\
%=&\left(\mathbf{u}_{t}-\mathbf{u}_{t-1}\right)\hat{\mathbf{W}}^{\top}\hat{\mathbf{W}}\mathbf{u}_{t}^{\top}\\
%=&\Delta\mathbf{u}_{t}\hat{\mathbf{W}}^{\top}\hat{\mathbf{W}}\mathbf{u}_{t}^{\top}
%\end{split}
%\end{equation}
%The gradient of the power consumption in the surrogate network with respect to the layer's weight matrix is 
%\begin{equation}
%\frac{\partial J_{P}}{\partial\hat{\mathbf{W}}}=(\hat{P}-P)\hat{\mathbf{W}}\left(\Delta\mathbf{u}_{t}^{\top}\mathbf{u}_{t}+\mathbf{u}_{t}^{\top}\Delta\mathbf{u}_{t}\right)

\section*{Acknowledgements}

This material is based on research sponsored by the Air Force Research Laboratory under agreement number FA8750-20-2-0503. The U.S. Government is authorized to reproduce and distribute reprints for Governmental purposes notwithstanding any copyright notation hereon.  The views and conclusions contained herein are those of the authors and should not be interpreted as necessarily representing the official policies or endorsements, either expressed or implied, of the Air Force Research Laboratory or the U.S. Government.

\bibliographystyle{splncs04}
\bibliography{refs.bib}

\begin{thebibliography}{10}
\providecommand{\url}[1]{\texttt{#1}}
\providecommand{\urlprefix}{URL }
\providecommand{\doi}[1]{https://doi.org/#1}

\bibitem{mnist}
{The MNIST database}. \url{http://yann.lecun.com/exdb/mnist/}

\bibitem{sc_net}
Batina, L., Bhasin, S., Jap, D., Picek, S.: Csi neural network: Using
  side-channels to recover your artificial neural network information (October
  2018)

\bibitem{biggio2018wild}
Biggio, B., Roli, F.: Wild patterns: Ten years after the rise of adversarial
  machine learning. Pattern Recognition  \textbf{84},  317--331 (2018)

\bibitem{amlpapers}
Carlini, N.: {A Complete List of All (arXiv) Adversarial Example Papers}.
  \url{https://nicholas.carlini.com/writing/2019/all-adversarial-example-papers.html}

\bibitem{dalvi2004adversarial}
Dalvi, N., Domingos, P., Sanghai, S., Verma, D.: Adversarial classification.
  In: Proceedings of the tenth ACM SIGKDD international conference on Knowledge
  discovery and data mining. pp. 99--108 (2004)

\bibitem{fgsm}
Goodfellow, I., Shlens, J., Szegedy, C.: Explaining and harnessing adversarial
  examples (2015)

\bibitem{cnn_extract}
Hua, W., Zhang, Z., Suh, G.: Reverse engineering convolutional neural networks
  through side-channel information leaks (November 2018)

\bibitem{model_extract}
Jagielski, M., Carlini, N., Bethelot, D., Kurakin, A., Papernot, N.: High
  accuracy and high fidelity extraction of neural networks (March 2020)

\bibitem{joseph2018adversarial}
Joseph, A.D., Nelson, B., Rubinstein, B.I., Tygar, J.: Adversarial Machine
  Learning. Cambridge University Press (2018)

\bibitem{lowd2005adversarial}
Lowd, D., Meek, C.: Adversarial learning. In: Proceedings of the eleventh ACM
  SIGKDD international conference on Knowledge discovery in data mining. pp.
  641--647 (2005)

\bibitem{lowd2005good}
Lowd, D., Meek, C.: Good word attacks on statistical spam filters. In: CEAS.
  vol.~2005 (2005)

\bibitem{bb_ml}
Papernot, N., McDaniel, P., Goodfellow, I., Somesh, J., Berkay~Celik, Z.,
  Swami, A.: Practical black-box attacks against machine learning (March 2017)

\bibitem{szegedy2013intriguing}
Szegedy, C., Zaremba, W., Sutskever, I., Bruna, J., Erhan, D., Goodfellow, I.,
  Fergus, R.: Intriguing properties of neural networks. arXiv preprint
  arXiv:1312.6199  (2013)

\bibitem{model_data}
Troung, J., Maini, P., Walls, R., Papernot, N.: Data-free model extraction
  (November 2020)

\bibitem{vorobeychik2018adversarial}
Vorobeychik, Y., Kantarcioglu, M.: Adversarial machine learning. Synthesis
  Lectures on Artificial Intelligence and Machine Learning  \textbf{12}(3),
  1--169 (2018)

\bibitem{sc_trace}
Wei, L., Liu, Y., Luo, B., Xu, Q.: I know what you see: Power side-channel
  attack on convolutional neural network accelerators (March 2018)

\bibitem{westecircuits}
Weste, N., Harris, D., CMOS~VLSI, D.: CMOS VLSI Design: A Circuits and Systems
  Perspective. Addison-Wesley, Reading, MA (2005)

\bibitem{yang2019quantization}
Yang, J., Shen, X., Xing, J., Tian, X., Li, H., Deng, B., Huang, J., Hua, X.s.:
  Quantization networks. In: Proceedings of the IEEE/CVF Conference on Computer
  Vision and Pattern Recognition. pp. 7308--7316 (2019)

\bibitem{model_fpga}
Yoshida, K., Kubota, T., Shiozaki, M., Fujino, T.: Model-extraction attack
  against fpga-dnn accelerator utilizing correlation electromagnetic analysis
  (2019)

\bibitem{sc_def}
Zhou, Y., Feng, D.: Side-channel attacks: Ten years after its publication and
  the impacts on cryptographic module security testing (2005)

\end{thebibliography}

\end{document}